\documentclass[sigconf]{acmart}
\renewcommand\footnotetextcopyrightpermission[1]{} %remove copyright
\usepackage{subcaption}
\usepackage{multirow}
\usepackage{enumitem}
\copyrightyear{2026}
\acmYear{2026}
\setcopyright{cc}
\setcctype{by}
\acmConference[MM '26] {Proceedings of the 34th ACM International Conference on Multimedia}{November 10--14, 2026}{Rio de Janeiro, Brazil.}
\acmBooktitle{Proceedings of the 34th ACM International Conference on Multimedia (MM '26), November 10--14, 2026, Rio de Janeiro, Brazil}
\acmISBN{979-8-4007-2213-4/2026/11}
\acmDOI{10.1145/3767308.3836513}

\begin{document}

\title{Let Prompts Bridge Defense Knowledge: Transferable Graph Purification via Vulnerability-Aware GPL}

\author{Shuomin Xue}
\authornote{Equal contribution.}
\orcid{0009-0005-8988-4221}
\affiliation{%
  \institution{Southeast University}
  \city{Nanjing}
  \country{China}
}
\email{shuominxue@seu.edu.cn}

\author{Jingyuan Li}
\authornotemark[1]
\affiliation{%
  \institution{Southeast University}
  \city{Nanjing}
  \country{China}
}
\email{lijingyuan@seu.edu.cn}

\author{Ju Jia}
\authornote{Corresponding author.}
\affiliation{%
  \institution{Southeast University}
  \city{Nanjing}
  \country{China}
}
\email{jiaju@seu.edu.cn}

\author{Jingxuan Yu}
\affiliation{%
  \institution{Southeast University}
  \city{Nanjing}
  \country{China}
}
\email{yujingxuan24@seu.edu.cn}

\author{Xiaojun Jia}
\affiliation{%
  \institution{Nanyang Technological University}
  \city{Singapore}
  \country{Singapore}}
  \email{jiaxiaojunqaq@gmail.com}

\begin{abstract}
Graph Neural Networks (GNNs) have emerged as a cornerstone for representing complex relational dependencies in diverse multimedia tasks, particularly in cross-platform user interest modeling and cross-modal semantic alignment.
In the real world, a practical defense against graph adversarial perturbations is needed. However, we observe that the prevailing adversarial purification methods are essentially domain-restricted defenses, which leads to the following shortcomings: (1) single-domain data provides insufficient structural and semantic diversity for learning robust purification criteria; (2) training of domain-specific defense strategies from scratch consumes substantial computational cost. 
To address the above limitations, we propose a transferable graph purification scheme, named ProGAP, to bridge adversarial defense knowledge via vulnerability-aware graph prompt learning. Firstly, to capture universal adversarial patterns, a perturbation-capture edge detector is pretrained on data-rich graphs by jointly modeling topological and semantic information. Subsequently, to achieve more knowledge transfer w.r.t. robustness, vulnerability-aware prompts are designed that inject targeted purification guidance into biased nodes, during which the pretrained detector adapts to distribution shifts in downstream graphs without parameter-laborious updates. 
Experimental results demonstrate that compared with state-of-the-art baselines, our ProGAP achieves 1\%-9\% improvement, and reduces the time consumption by up to 2.2x. The code for ProGAP is available at https://github.com/Lieyoufffff/ProGAP.
\end{abstract}




\maketitle

\section{Introduction}
\begin{figure}[t]
    \centering
    \includegraphics[width=0.48\textwidth]{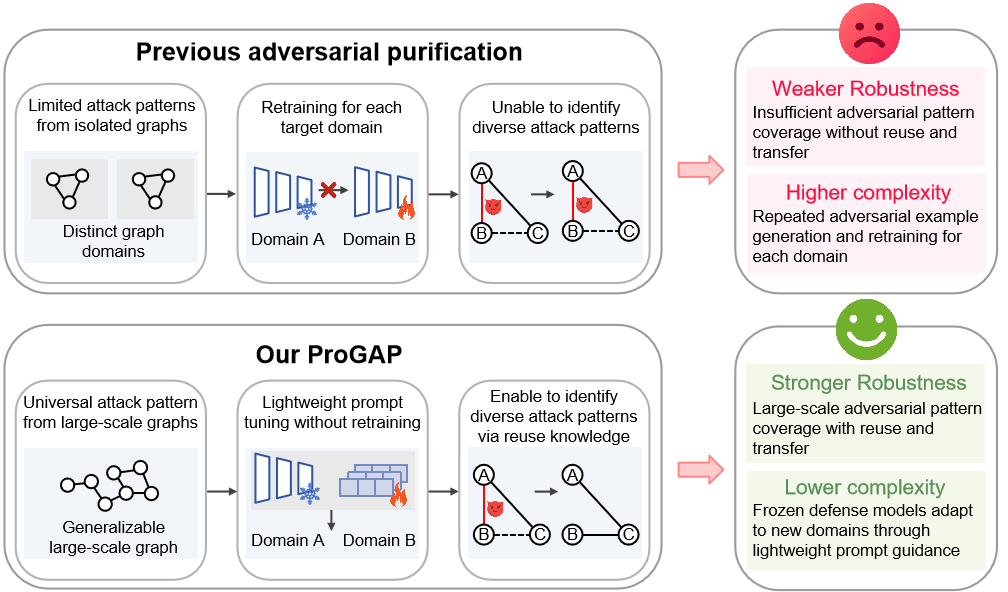}
    \caption{The comparison between our scheme and other
adversarial purification frameworks.}
    \label{fig:intro}
\end{figure}
Graph Neural Networks (GNNs) have been widely applied to structured multimedia data~\cite{multimedia1,multimedia2,yuan2026promptguard,fang2026disentangling,DBLP:conf/coling/PeiYZCJ25}, which includes cross-modal event extraction from heterogeneous media streams and personalized multimedia recommendation based on user content interaction graphs~\cite{BridgeGLM, multimedia3}. The topology awareness of GNNs originates from their message propagation mechanism~\cite{graphsage,yu2026mpas}, in which node features and hidden representations are treated as messages and propagated within the k-ego subgraph.

The reliance of GNNs on the underlying graph structure makes them extremely vulnerable to adversarial attacks, as even minor perturbations can severely compromise their performance~\cite{metatack,ctea,metacon,tan2026practical,DBLP:journals/tifs/WuCLXJHFLX25}. Therefore, a reliable defense strategy is needed. Among various defense strategies, adversarial purification serves as a mainstream approach~\cite{STABLE,CASR,gprgae}, which trains a specialized purifier to remove adversarial edges before message passing. Compared with alternative defense methods, adversarial purification offers a more effective solution by eliminating adversarial perturbations directly on the graph. Most of the previous graph purification methods either rely on graph attributes to detect adversarial edges~\cite{gcnsvd}, or generate adversarial examples by perturbing clean graphs to train edge detectors~\cite{gprgae}.

However, most of the previous methods are limited to single-domain defense and fail to generalize across diverse multimedia scenarios where graph-structured data exhibit significant distribution shifts, which raises two limitations.
First, the single-domain defense leads to insufficient coverage against adversarial knowledge. Adversarial perturbations typically exploit vulnerable graph structures, which appear with limited frequency within a single domain~\cite{DBLP:conf/kdd/JiaY00Z025}. Therefore, this insufficient coverage of adversarial knowledge prevents purifiers from establishing robust criteria to distinguish adversarial perturbations from normal structural changes, which leads to sub-optimal performances.
Second, they are inherently domain-constrained operations, i.e., should be re-trained across different tasks. This involves repeated adversarial example generation, which incurs substantial computational consumption.

In this work, to address the limited cross-domain transferability of existing adversarial purification methods for graph-structured multimedia data, we propose a prompt-guided transferable adversarial purification
scheme, named ProGAP. Instead of training a separate defense model for each target graph, our scheme enables the reuse of adversarial knowledge across multimedia domains through prompt guidance, thereby improving deployment efficiency.
 Specifically, to capture general adversarial edge patterns, we first pretrain a perturbation-capture edge detector on the upstream graph by jointly modeling structural and semantic information. Secondly, to ensure knowledge transfer w.r.t. robustness, vulnerability-aware prompts are introduced to encode perturbation-sensitive information in downstream graphs, which provide adaptive guidance for the frozen edge detector when graph distributions shift under different attacks. In summary, our contributions are as follows:

\begin{itemize}[leftmargin=*, labelsep=0.4em]
\setlength{\itemsep}{0pt}
\setlength{\parskip}{0pt}
\setlength{\topsep}{0pt}
\setlength{\parsep}{0pt}
\item \textbf{Shortcoming identification.} We reveal that prevailing adversarial purification methods are confined to single-domain operation, which limits the coverage of vulnerability patterns required for learning robust purification criteria, and incurs substantial computational cost due to domain-specific retraining. Consequently, they exhibit low transferability and scalability when deployed in large-scale, cross-domain multimedia graph scenarios.
\item \textbf{Advanced solution.} 
We introduce our scheme, named ProGAP, which is a vulnerability-aware adversarial purification scheme via graph prompt learning. To learn general adversarial edge characteristics, a perturbation-capture edge detector is trained on a data-rich graph that integrates structural and semantic information upstream. Furthermore, to achieve more knowledge transfer w.r.t. robustness, vulnerability-aware prompts are designed for downstream biased nodes that are vulnerable to attacks, which guide the purification process and enable adaptation to domain shifts.
\item \textbf{Comprehensive evaluation.} We conducted extensive experiments across various graph-structured media datasets. 
To demonstrate the superiority of knowledge transfer for adversarial defenses, we investigate the use of vulnerability-targeted prompts. Experimental results reveal that, compared with eight state-of-the-art (SOTA) baselines, ProGAP outperforms them in 78\% of the evaluated scenarios and reduces the running time on PubMed from 22.85s to 10.11s.

\end{itemize}

\section{Related Work}
\subsection{Graph Adversarial Defense}
Numerous attempts have been made to enhance the robustness of GNNs~\cite{DBLP:journals/tdsc/JiaLWMWD25}. Among these defense strategies, purification is a relatively promising solution for downstream applications, as it trains a purifier to preprocess input data. GCN-SVD \cite{gcnsvd} performs a low-rank approximation to purify the graph. GPR-GAE \cite{gprgae} separates robustness from the classifier and leverages multiple Generalized PageRank filters for graph purification. Some methods forgo adversarial purification and focus on designing a robust GNN architecture as a defense. ProGNN \cite{prognn} jointly learns the low-rank, sparsity, and feature smoothness properties of the graph, while GADC \cite{gadc} proposes a min-max optimization formulation based on Laplacian distance perturbations to achieve graph adversarial diffusion convolutions. However, their methods are limited to a single domain and cannot transfer across domains.
\subsection{Graph Knowledge Transfer}
Graph knowledge transfer focuses on the cross-domain adaptation of structural priors and representational patterns learned from source graphs to enhance generalization on target domains~\cite{kooverjee2022investigatingtransferlearninggraph,graphlora}. Among them, pre-training and fine-tuning has emerged as the predominant paradigm, in which a GNN is pretrained on the source graph and subsequently fine-tuned on the target graph \cite{transfer1,graphcontrol,DBLP:journals/tifs/JiaLWFMWD25}. Despite its success, full fine-tuning is frequently inefficient and susceptible to challenges such as overfitting and catastrophic forgetting \cite{transfer2,transfer3}. To reduce resource requirements, graph prompt learning has been increasingly adopted as a lightweight mechanism for knowledge transfer. 

\subsection{Cross-Domain Graph Prompt Learning}
In recent two years, many graph prompt paradigms have been proposed, which aim to extract key information from prior knowledge through token-wise guidance for downstream tasks~\cite{uniprompt,PAGPL/AAAI26}.  
GPPT \cite{gppt} modifies the standalone node into a token pair. GPF~/~GPF-Plus \cite{GPF} proposes a universal prompt tuning method by introducing additional learnable parameters as a prompt in the feature space of the input graph. All-in-One \cite{all-in-one} extends the standalone token to a prompt graph. MultiGprompt~\cite{yu2024multigprompt} introduces a multi-task pre-training and dual-prompt mechanism to aggregate comprehensive knowledge from multiple pretext tasks. However, all of these methods focus on the compatibility between pre-training and downstream tasks, while we utilize vulnerability-aware prompts to enable cross-domain transfer of defensive knowledge on graphs.

\section{Preliminary}
\subsection{Notations}
Let $\mathcal{G} = \{ \mathcal{V}, \mathcal{E} \}$ represent an undirected and unweighted graph comprising $N$ nodes. Here, $\mathcal{V}$ and $\mathcal{E}$ are the sets of nodes and edges, respectively. The adjacency matrix is defined as $A \in \{0,1\}^{n \times n}$, where $a_{ij}=1$ indicates that node $v_{i}$ connects to node $v_j$, and $a_{ij}$ = 0 represents no connection. The original features of all nodes are summarized as a matrix $X\in R^{N\times d}$, and $x_{i}$ indicates the feature of node $v_{i}$. Moreover, the label of all nodes are denoded as $\mathbf{Y}$, each node has a label \( y_i \in C \), where \( C = \{c_1, c_2, \dots, c_K\} \). 

The set of labeled nodes is written as \(\mathcal{V}_L \subseteq \mathcal{V}\) and the unlabeled nodes are denoted by \(\mathcal{V}_U = \mathcal{V}/\mathcal{V}_L\). We denote the pretrained GNN purifier by \(\psi^*\) and introduce a set of learnable prompt tokens $\boldsymbol{P}=\{p_1,...,p_N\}$, which are injected into the graph through a prompting function \(\mathcal{P}\). \(\mathcal{P}\) maps the attacked graph inputs to a modified representation, written as \(\mathcal{P}(X, A, \boldsymbol{P})\). The prediction loss associated with the downstream task is referred to as $\mathcal{L}_{\text{Task}}(\cdot)$
.

\subsection{Threat Models}

In this paper, we adopt two attack scenarios: the commonly used gray-box global poisoning attack and the white-box adaptive attack \cite{metatack,grad}. We focus on adversarial attack through structural perturbations, where attackers manipulate edges (insertions or deletions) to degrade performance. 
In the setting of the former scenario, attackers have visibility into the graph data and labels. The attacker's objective is to find an optimal perturbed graph $\hat{\mathcal{G}}$ that maximally impairs the overall performance of the downstream classifier, which can be formulated as follows \cite{poison_attack2021robustness}:
\begin{align}
argmin_{\hat{A} \in \Phi(A)} \mathcal{L}_{atk}(f_{\theta^*}(\hat{A}, X), y),
\end{align}
where $\hat{A}$ is the perturbed adjacency matrix, and $\Phi(A)$ is a set of adjacency matrices that satisfies: $\frac{\|\hat{A} - A\|_0}{\|A\|_0} \leq \Delta$, in which $\Delta$ is the maximum perturbation rate. The $\mathcal{L}_{atk}$ can be computed using ground-truth labels of training nodes or labels (or pseudo-labels) of unlabeled nodes. The $\theta^*$ refers to the parameters of the surrogate GNN, which simulates the real-world scenarios for the precise modification of attacks. 

Adaptive attacks are formed in a white-box setting, where the attacker possesses complete information, including features, graph structure, labels, and all details of the defender’s model \cite{adaptive_attack}. For instance, w.r.t. our scheme, the attacker has complete access to the pretrained edge detector, the vulnerability-targeted prompts $\boldsymbol{P}=\{p_1,...,p_N\}$, and the downstream classifier. The attacker's objective is to generate perturbations capable of bypassing prompt-guided purification processes while maximizing classification loss after purification. 

\begin{figure*}[t]
    \centering
    \includegraphics[width=1.00\textwidth]{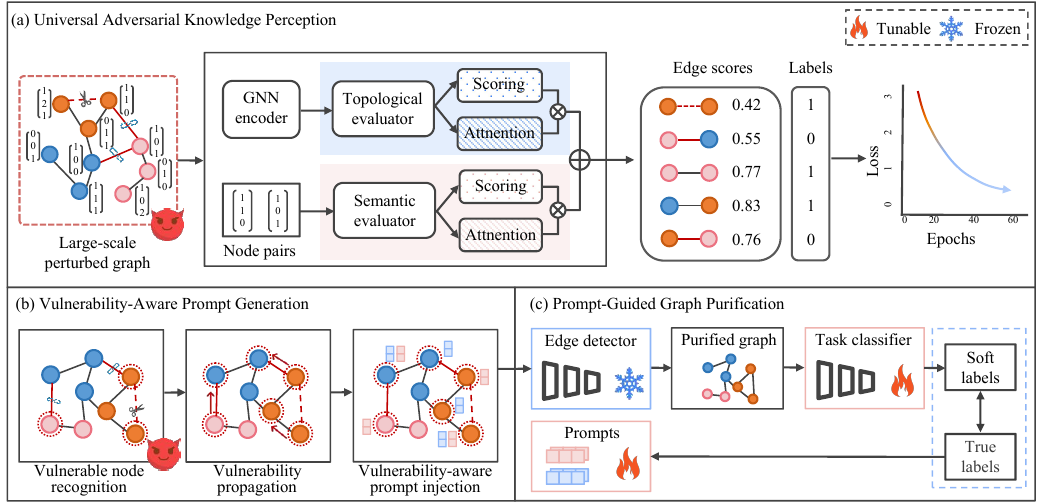}
    \caption{Our scheme consists of three modules, including universal adversarial knowledge perception, vulnerability-aware prompt generation, and prompt-guided graph purification.}
    \label{fig:overall}
\end{figure*}
\section{Method}
\subsection{Overview}
As shown in Figure~\ref{fig:overall}, ProGAP is presented as a lightweight framework for cross-domain graph purification, which can rapidly adapt to previously unseen target graphs and generate effective purification. The scheme under discussion comprises the following three constituent parts.

First, a malicious perturbation capturer is trained to decouple structural and semantic perturbations in the source domain, which builds the topological structure and feature space to capture the general representation of vulnerability perturbations. Subsequently, in target domains, prompts designed for purification are constructed to reveal vulnerable nodes susceptible to adversarial attacks. Eventually, guided by the prompts, the pretrained malicious perturbation capturer continuously optimizes the detection of attack edges, and the purified result is ultimately treated as a clean graph for real-world detection.
\subsection{Universal Adversarial Knowledge Perception}

To learn a universal robust representation, a perturbation-capture edge detector is pretrained that disentangles structural and semantic fluctuations.
First, the clean graph is perturbed using PGD \cite{xu2019topologyattackdefensegraph}, as it is a representative gradient-based attack that exposes universal adversarial patterns, to obtain the perturbed graph $\hat{\mathcal{G}}$, and the binary label for each edge between node pairs $(u,v)$ is defined as follows:
\begin{align}
\ell_{uv}=
\begin{cases}
0, & (u,v)\in \hat{\mathcal{E}}\ \text{and}\ (u,v)\notin \mathcal{E},\\[4pt]
1, & (u,v)\in \mathcal{E}.
\end{cases}
\end{align}
Here, $\hat{\mathcal{E}}$ represents the perturbed edge set, label 0 represents the edge inserted by the attacker, and label 1 represents a preserved normal edge that existed in the original clean graph.

The edge detector consists of two complementary modules: 
the topological evaluator ($D_{t}$) and the semantic evaluator ($D_{s}$). 
$D_{t}$ employs a GCN to learn structure-aware node representations. To prevent the unreliability of these representations due to structural perturbations, the learned embeddings with original node features are concatenated:
\begin{equation}
\mathbf{e}_{uv}=\mathrm{CONCAT}[f_{\theta}(\hat{\mathcal{G}})_{u},x_{u},f_{\theta}(\hat{\mathcal{G}})_{v},x_v],
\end{equation}
where $\theta$ denotes the GCN parameters. These embeddings are then fed into a topology evaluator consisting of linear layers that output a score $t_{uv}$, which indicates the probability of edge $(u,v)$ being adversarial. 

$D_{s}$ calculates semantic similarity between nodes based on the proximity of node pairs. Specifically, we compute the cosine similarity between two nodes to infer potential edge connections:
\begin{equation}
s_{uv} = \frac{X_u \odot X_v}{||X_u||_2||X_v||_2}.
\label{eq:similarity}
\end{equation}
To adaptively fuse information from both modules, we introduce learnable attention weights that measure the relative importance of structural and semantic information. The final edge score $d_{uv}$ is computed as:
\begin{gather}
    d_{uv} = \alpha_{1} \cdot t_{uv} + (1-\alpha_{1}) \cdot s_{uv},
\end{gather}
where $\alpha_{1}$ is a learnable weight to balance the relative importance of structural and semantic information, which enables the detector to have flexible adaptation to diverse graph distributions and graph structures. 
The edge detector is optimized by minimizing the cross-entropy loss between predicted edge scores $d_{uv}$ and ground truth labels $\ell_{uv}$:
\begin{equation}
\mathcal{L}_{pre} = -\sum_{(u,v)\in \mathcal{E}\cup\hat{\mathcal{E}}}
\Big[ \ell_{uv}\log d_{uv} + (1-\ell_{uv})\log(1-d_{uv}) \Big].
\end{equation}

\subsection{Vulnerability-Aware Prompt Generation}
It is observed that even when the downstream data closely matches that of the pre-training tasks, simple deceptive techniques can still generate highly misleading prompts. Adversarial perturbations exhibit node-level heterogeneity, which tends to concentrate on nodes with specific structural or semantic vulnerabilities. The uniform application of prompts across all nodes fails to align defensive guidance with vulnerable nodes, thereby weakening the effectiveness of defense transfer.

To achieve more robust transfer, we inject vulnerability-aware prompts into biased nodes. It is observed that adversarial attacks typically exploit two types of nodes \cite{gpromprshield}: (1) low-degree nodes are inherently vulnerable due to their limited connectivity and reduced neighborhood information, making them less stable compared to high-degree nodes with stronger community characteristics; (2) most adversarial attacks exploit graph homophily by introducing heterophily through dissimilar connections. Nodes with low similarity to their neighbors become prime attack targets, as their feature consistency can be easily disrupted. Therefore, we first identify
two types of sets of vulnerable nodes: low-degree nodes and low-similarity nodes. To make the identification process adaptive across
graphs with different structural properties, we first estimate the
global homophily level of the graph as:

\begin{equation}
H = \frac{1}{|\mathcal{E}|} \sum_{(u,v) \in \mathcal{E}} \mathrm{sim}(X_u, X_v)
\end{equation}
where $\mathrm{sim}(\cdot, \cdot)$ denotes the cosine similarity between node features. The identification of vulnerable nodes is categorized into two dimensions:
\begin{gather}
S_{d} = \left\{ u \mid |\mathcal{N}_u| < \tau_{\text{degree}} \right\}, \quad \tau_{\text{degree}} = \mu_d - H \cdot \sigma_d, \\
S_{s} = \left\{ u \mid \frac{1}{|\mathcal{N}_u|} \sum_{v \in \mathcal{N}_u} sim(X_u, X_v) < \tau_{sim} \right\}, \quad \tau_{sim} = \mu_s - H \cdot \sigma_s,
\end{gather}
where $\mathcal{N}_{u}$ denotes the set of neighbors of node $u$, $\tau_{\text{degree}}$ is the degree threshold. The terms $\mu_{d}$ and $\mu_{s}$ denote the mean node degree and mean neighborhood similarity, respectively, while $\sigma_{d}$ and $\sigma_{s}$ represent the standard deviation of the node degree distribution and the neighborhood similarity distribution, respectively.

Given that the GNN progressively aggregates and updates node-level messages from k-ego graphs via a message passing mechanism, adversarial effects can propagate from vulnerable nodes to their neighbors. 
To comprehensively identify all potentially biased nodes, a connection score is calculated.
Specifically, the connection score is defined as the proportion of the number of edges connected to normal nodes to the total number of edges:
\begin{align}
h_{d}^{u} &=
\frac{
\left|\mathcal{E}_{u \leftrightarrow \overline{S_{d}}}\right|
}{
\left|\mathcal{E}_{u \leftrightarrow S_{d}}\right|
+
\left|\mathcal{E}_{u \leftrightarrow \overline{S_{d}}}\right|
}, 
\quad
S'_{d} = \{u \mid h_{d}^{u} < \tau_{l}^{d}\},
\end{align}
\begin{align}
h_{s}^{u} &=
\frac{
\left|\mathcal{E}_{u \leftrightarrow \overline{S_{s}}}\right|
}{
\left|\mathcal{E}_{u \leftrightarrow S_{s}}\right|
+
\left|\mathcal{E}_{u \leftrightarrow \overline{S_{s}}}\right|
},
\quad
S'_{s} = \{u \mid h_{s}^{u} < \tau_{l}^{s}\},
\end{align}
where $\mathcal{E}_{\{u\} \leftrightarrow S}$ denotes the set of edges connecting node $u$ to nodes in set $S$, and $\overline{S_d}$ and $\overline{S_s}$ denote the complements of the low-degree node set $S_d$ and the low-similarity node set $S_s$, respectively. A lower connection score indicates that a node is predominantly connected to vulnerable nodes rather than normal nodes. Therefore, nodes whose connection scores fall below thresholds $\tau_l^d$ and $\tau_l^s$ are identified as biased nodes.

Two learnable defensive prompts are directly added to the feature space of nodes: a degree vulnerability prompt token $p_d \in \mathbb{R}^d$ for nodes in $S_d'$ and a similarity vulnerability prompt token $p_s \in \mathbb{R}^d$ for nodes in $S_s'$, where $d$ is the feature dimension:
\begin{equation}
X^*[S_{d}'] = X[S'_{d}] + p_d,\quad X^*[S'_{s}] = X[S'_{s}] + p_s.
\end{equation}
Since a node may belong to both $S'_d$ and $S'_s$ simultaneously, the actual prompt for each node is selected based on the set to which it belongs from $\{(), (p_d), (p_s), (p_d, p_s)\}$. To adaptively balance multiple prompts, we introduce learnable fusion weights. The final vulnerability-aware prompt for node $u$ is computed as follows:
\begin{align}
X^{*}[u] &= X[u] + \sum_{n=1}^{i} w_n p_n,
\end{align}
where $i$ denotes the number of prompts assigned to node $u$, and $w_n$ denotes the fusion weight of prompt $p_n$, which can balance different defensive prompts according to each node's vulnerability profile.
\subsection{Prompt-Guided Graph Purification}

To enable adaptive cross-domain transfer, the purification process is reformulated as a context-guided co-optimization of feature tokens and graph topology. Specifically, the prompt-enhanced graph is fed into two pretrained edge evaluation modules to obtain the final edge score $d_{uv}'$. Edges with $d_{uv}'$ below the decision threshold $\tau_{edge}$ are identified as adversarial edges and removed. By applying this process to all edges, we obtain the purified graph $ \mathcal{G}_{p} $.

Given that adversarial perturbations often compromise local structural integrity while leaving global feature distributions relatively stable \cite{Evennet}, we introduce a structure-agnostic MLP to anchor the optimization. By leveraging global feature consistency, the MLP generates high-confidence pseudo-labels $\hat{y}_i$ for a small set of unlabeled nodes $\mathcal{V}_{pseudo}$ and extends them to the training set,thereby constructing an augmented training set $\mathcal{V}_{aug}$. This effectively provides a robust semantic prior that bypasses unreliable neighborhood dependencies: 

\begin{equation}
\mathcal{V}_{aug} = \mathcal{V}_{L} \cup \mathcal{V}_{pseudo}.
\end{equation}
Consequently, our scheme establishes a coupled optimization chain where gradients from the classification loss $\mathcal{L}_{cls}$ are backpropagated to the vulnerability-aware prompts $\{p_d, p_s\}$ and the classification head $\theta_{1}$, simultaneously refining the feature augmentations and the purified topology:
\begin{equation}
\min_{\theta_{1},\, p_d,\, p_s} \; \mathcal{L}_{cls}
= -\sum_{u_{i} \in \mathcal{V}_{aug}} y_i \log\big(f_{\theta_{1}}(\mathcal{G}_{p})\big),
\end{equation}
where $y_i$ denotes the true label~(pseudo-labels) of node $u_{i}$. 

\begin{table*}[tb]
\centering
\caption{The accuracy (in $\%\ \pm\ \sigma$) comparison between our scheme and baselines under clean and attacked settings (25\% and 50\% perturbation). M, G, and H denote Metattack, GraD, and Heuristic attacks, respectively. ProGAP-Pu, ProGAP-C, and ProGAP-Ph refer to using PubMed, Computers, and Photo as upstream pre-training datasets, respectively.}
\label{tab:performance_node_task}
\setlength{\tabcolsep}{2pt}
\small
\begin{tabular}{c c|cccccccccc}
\toprule
Dataset & Attack Rate & GCN & GCN-SVD & JaccardGCN & GNNGuard & HANG-quad & GADC & GPR-GAE & ProGAP-Pu & ProGAP-C & ProGAP-Ph\\
\midrule

\multirow{7}{*}{Cora} & Clean & 82.48$\pm$0.39 & 77.78$\pm$0.98 & 80.90$\pm$1.26 & \textbf{82.52}$\pm$\textbf{1.09} & 81.06$\pm$2.26 & 79.57$\pm$0.42 & 66.14$\pm$1.10 & 80.49$\pm$1.43 & 79.78$\pm$0.66 & 77.53$\pm$0.52 \\
& M-25 & 58.82$\pm$4.39 & 63.86$\pm$4.38 & 64.60$\pm$0.61 & 57.99$\pm$4.08 & 63.18$\pm$1.29 & 73.70$\pm$1.02 & 65.37$\pm$1.03 & 77.41$\pm$0.89 & \textbf{77.73}$\pm$\textbf{0.37} & 74.69$\pm$1.37 \\
& M-50 & 35.23$\pm$1.16 & 44.13$\pm$5.39 & 48.32$\pm$0.96 & 35.70$\pm$4.45 & 62.29$\pm$1.34 & 68.15$\pm$0.76 & 65.20$\pm$1.18 & \textbf{77.31}$\pm$\textbf{1.04} & 75.21$\pm$1.96 & 73.32$\pm$1.10 \\
& H-25 & 71.97$\pm$2.27 & 68.95$\pm$1.00 & 71.60$\pm$2.26 & 71.39$\pm$1.90 & 63.46$\pm$0.93 & 75.57$\pm$0.50 & 65.74$\pm$1.82 & 77.13$\pm$1.16 & \textbf{77.41}$\pm$\textbf{0.60} & 74.72$\pm$0.98 \\
& H-50 & 63.53$\pm$2.63 & 55.91$\pm$2.46 & 68.59$\pm$2.58 & 64.96$\pm$3.01 & 63.76$\pm$0.69 & 71.07$\pm$1.32 & 66.46$\pm$1.53 & \textbf{75.63}$\pm$\textbf{1.24} & 75.52$\pm$1.81 & 72.39$\pm$0.19 \\
& G-25 & 53.98$\pm$4.02 & 66.22$\pm$1.93 & 60.59$\pm$2.58 & 53.97$\pm$3.07 & 63.08$\pm$0.64 & 72.34$\pm$0.89 & 66.14$\pm$1.25 & \textbf{76.94}$\pm$\textbf{1.39} & 76.91$\pm$0.91 & 73.75$\pm$1.03 \\
& G-50 & 30.61$\pm$4.92 & 52.91$\pm$3.44 & 51.78$\pm$2.46 & 33.98$\pm$4.31 & 62.64$\pm$0.08 & 68.94$\pm$0.77 & 65.59$\pm$3.53 & \textbf{74.96}$\pm$\textbf{1.33} & 73.97$\pm$1.12 & 72.97$\pm$0.28 \\
\midrule
\multirow{7}{*}{CiteSeer} & Clean & 73.20$\pm$0.53 & 68.63$\pm$1.64 & \textbf{73.67}$\pm$\textbf{0.60} & 73.66$\pm$1.16 & 63.17$\pm$0.18 & 72.75$\pm$0.90 & 65.15$\pm$2.04 & 73.15$\pm$0.77 & 73.15$\pm$0.49 & 70.73$\pm$1.34 \\
& M-25 & 57.52$\pm$1.88 & 61.54$\pm$1.03 & 61.42$\pm$1.61 & 57.56$\pm$4.36 & 63.72$\pm$0.81 & 68.89$\pm$0.27 & 65.52$\pm$1.17 & \textbf{70.92}$\pm$\textbf{1.22} & 70.65$\pm$0.71 & 68.35$\pm$0.89 \\
& M-50 & 41.15$\pm$3.86 & 52.03$\pm$1.34 & 48.72$\pm$3.96 & 43.83$\pm$4.26 & 61.93$\pm$1.19 & 61.01$\pm$1.62 & 64.99$\pm$2.24 & 69.70$\pm$1.91 & \textbf{73.97}$\pm$\textbf{1.12} & 66.23$\pm$1.19 \\
& H-25 & 60.13$\pm$0.77 & 58.66$\pm$1.83 & 61.61$\pm$0.78 & 60.83$\pm$1.13 & 63.72$\pm$0.70 & 67.38$\pm$1.26 & 64.97$\pm$1.03 & 69.82$\pm$1.51 & \textbf{71.13}$\pm$\textbf{1.37} & 68.36$\pm$1.59 \\
& H-50 & 52.99$\pm$0.47 & 49.80$\pm$1.85 & 57.87$\pm$1.49 & 55.95$\pm$0.92 & 63.70$\pm$0.36 & 63.34$\pm$1.55 & 64.74$\pm$1.22 & 69.50$\pm$1.20 & \textbf{69.94}$\pm$\textbf{0.93} & 67.02$\pm$1.25 \\
& G-25 & 53.77$\pm$2.62 & 63.07$\pm$1.09 & 57.83$\pm$2.57 & 53.87$\pm$1.60 & 63.80$\pm$0.58 & 65.88$\pm$1.29 & 65.62$\pm$1.18 & 70.05$\pm$1.49 & \textbf{70.53}$\pm$\textbf{1.00} & 67.91$\pm$1.83 \\
& G-50 & 38.34$\pm$3.80 & 56.07$\pm$3.77 & 47.04$\pm$1.33 & 40.44$\pm$2.75 & 62.66$\pm$0.27 & 60.18$\pm$3.70 & 64.83$\pm$1.38 & \textbf{68.72}$\pm$\textbf{1.68} & 68.56$\pm$1.31 & 65.21$\pm$1.17 \\
\midrule
\multirow{7}{*}{CoraML} & Clean & 83.94$\pm$0.57 & 81.33$\pm$0.75 & 84.52$\pm$0.58 & \textbf{85.38}$\pm$\textbf{0.45} & 73.21$\pm$0.05 & 78.93$\pm$0.63 & 74.76$\pm$3.56 & 82.01$\pm$0.60 & 81.41$\pm$1.01 & 82.13$\pm$0.67 \\
& M-25 & 60.02$\pm$3.11 & 74.74$\pm$0.90 & 62.56$\pm$2.84 & 59.42$\pm$3.89 & 73.04$\pm$0.38 & 77.04$\pm$0.68 & 72.20$\pm$1.16 & \textbf{78.62}$\pm$\textbf{1.21} & 78.28$\pm$1.20 & 78.15$\pm$1.35 \\
& M-50 & 38.86$\pm$1.60 & 46.37$\pm$0.25 & 41.85$\pm$1.19 & 41.43$\pm$1.97 & 72.84$\pm$0.46 & 69.87$\pm$0.98 & 72.13$\pm$2.57 & \textbf{77.26}$\pm$\textbf{0.72} & 76.18$\pm$1.49 & 74.32$\pm$1.01 \\
& H-25 & 66.37$\pm$1.97 & 75.75$\pm$0.61 & 71.92$\pm$1.64 & 73.20$\pm$1.67 & 73.01$\pm$0.45 & 76.53$\pm$0.84 & 72.34$\pm$2.59 & 79.35$\pm$0.30 & \textbf{79.44}$\pm$\textbf{0.76} & 79.04$\pm$0.61 \\
& H-50 & 66.84$\pm$1.68 & 64.88$\pm$1.11 & 65.05$\pm$2.19 & 65.69$\pm$1.59 & 73.13$\pm$0.61 & 73.47$\pm$0.46 & 72.18$\pm$0.78 & 77.51$\pm$0.36 & \textbf{79.88}$\pm$\textbf{0.66} & 77.02$\pm$0.91 \\
& G-25 & 50.11$\pm$2.93 & 68.35$\pm$1.17 & 54.87$\pm$3.81 & 72.03$\pm$1.11 & 73.26$\pm$0.19 & 75.53$\pm$0.78 & 72.84$\pm$1.16 & 79.12$\pm$0.51 & \textbf{79.44}$\pm$\textbf{0.53} & 78.74$\pm$0.75 \\
& G-50 & 29.75$\pm$3.68 & 43.63$\pm$7.00 & 35.38$\pm$5.22 & 34.45$\pm$3.93 & 71.93$\pm$0.22 & 71.92$\pm$0.78 & 72.29$\pm$2.81 & 75.94$\pm$2.00 & \textbf{78.51}$\pm$\textbf{2.09} & 76.98$\pm$1.09 \\
\midrule
\multirow{7}{*}{PubMed} & Clean & 83.96$\pm$0.14 & 84.67$\pm$0.06 & 86.53$\pm$0.34 & \textbf{86.63}$\pm$\textbf{0.20} & 84.63$\pm$0.09 & 82.87$\pm$0.24 & 84.91$\pm$0.27 & 85.28$\pm$0.25 & 84.60$\pm$0.49 & 85.37$\pm$1.47 \\
& M-25 & 70.53$\pm$1.30 & 82.62$\pm$0.32 & 70.59$\pm$1.50 & 70.59$\pm$1.50 & 77.17$\pm$2.86 & 75.40$\pm$4.03 & \textbf{83.97}$\pm$\textbf{0.27} & 83.64$\pm$0.51 & 83.39$\pm$0.16 & 83.62$\pm$0.61 \\
& M-50 & 65.57$\pm$1.01 & 81.23$\pm$0.34 & 67.43$\pm$1.43 & 68.53$\pm$1.35 & 75.17$\pm$2.32 & 73.20$\pm$1.24 & 81.31$\pm$0.11 & 81.98$\pm$0.66 & 76.18$\pm$1.49 & \textbf{82.01}$\pm$\textbf{0.29} \\
& H-25 & 78.83$\pm$1.42 & 82.91$\pm$0.30 & 79.89$\pm$1.10 & 79.54$\pm$1.39 & 80.26$\pm$2.87 & 77.20$\pm$2.00 & 82.73$\pm$0.18 & 83.21$\pm$0.51 & 83.19$\pm$0.52 & \textbf{84.06}$\pm$\textbf{0.30} \\
& H-50 & 63.53$\pm$2.63 & 81.62$\pm$0.22 & 72.75$\pm$2.23 & 73.34$\pm$2.37 & 77.25$\pm$2.85 & 52.59$\pm$1.33 & 82.22$\pm$0.37 & 84.08$\pm$0.57 & \textbf{84.27}$\pm$\textbf{0.63} & 82.98$\pm$0.75 \\
& G-25 & 74.90$\pm$0.43 & 81.43$\pm$0.57 & 80.07$\pm$0.84 & 79.84$\pm$0.43 & 79.19$\pm$2.87 & 76.66$\pm$3.94 & \textbf{82.99}$\pm$\textbf{0.98} & 82.31$\pm$0.57 & 82.05$\pm$0.48 & 82.04$\pm$0.47 \\
& G-50 & 76.34$\pm$0.96 & 81.28$\pm$0.46 & 78.69$\pm$2.01 & 78.67$\pm$2.11 & 77.04$\pm$2.68 & 74.58$\pm$2.59 & 79.51$\pm$0.88 & 79.06$\pm$1.14 & 78.59$\pm$1.25 & \textbf{81.39}$\pm$\textbf{1.09} \\
\bottomrule
\end{tabular}
\vskip -4mm
\end{table*}

\section{Experiment}

\begin{figure*}[!htbp]
    \centering
    \includegraphics[width=0.45\textwidth]{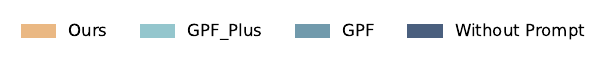}
    \label{fig:legend}
\end{figure*}

\begin{figure*}[!htbp]
    \centering
    \begin{subfigure}{0.245\textwidth}
        \centering
        \includegraphics[width=\linewidth]{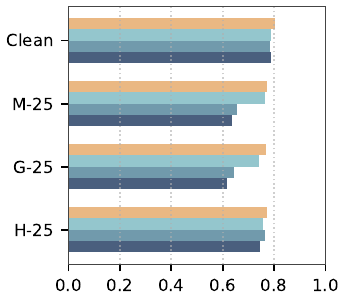}
        \caption{Cora}
        \label{fig:sub6}
    \end{subfigure}
    \hfill
    \begin{subfigure}{0.245\textwidth}
        \centering
        \includegraphics[width=\linewidth]{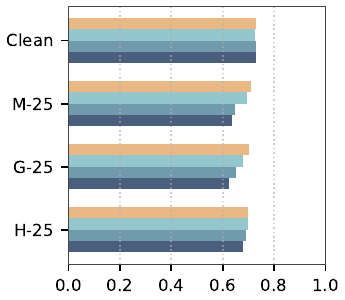}
        \caption{CiteSeer}
        \label{fig:sub7}
    \end{subfigure}
    \begin{subfigure}{0.245\textwidth}
        \centering
        \includegraphics[width=\linewidth]{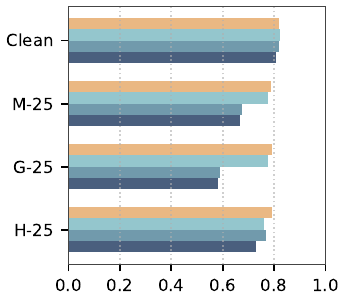}
        \caption{CoraML}
        \label{fig:sub8}
    \end{subfigure}
    \begin{subfigure}{0.245\textwidth}
        \centering
        \includegraphics[width=\linewidth]{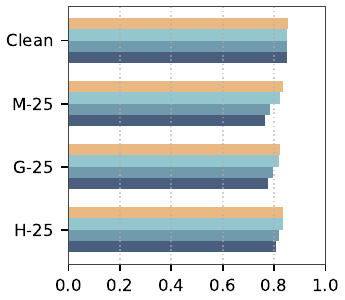}
        \caption{PubMed}
        \label{fig:sub9}
    \end{subfigure}
    \caption{The accuracy (in \%) comparison with and without prompt under different attacks at perturbation rates of 25\%.}
    \label{fig:prompt}
    \vskip -3mm
\end{figure*}

In this section, we conduct the experimental evaluation to address the following research questions:\\
\textbf{RQ1:}~Whether ProGAP outperforms SOTA defences when subjected to different types of non-adaptive adversarial attacks? \\
\textbf{RQ2:}~Does GPL enable effective and robust defense knowledge transfer to target domains? \\
\textbf{RQ3:}~Does ProGAP achieve higher efficiency than existing defense methods? \\
\textbf{RQ4:}~Can ProGAP maintain adversarial robustness in adaptive attack scenarios?\\
\textbf{RQ5:}~What is the impact of $D_{t}$, $D_{s}$, the attention mechanism, pseudo labels, and vulnerability-aware prompts of the proposed ProGAP?\\
\textbf{RQ6:}~Does ProGAP remain effective in scenarios involving transitions from homophily to heterophily graphs?\\
\subsection{Experimental Settings}
\textbf{Dataset.} We utilize ten benchmark graph-structured datasets for the evaluation, including
six homophilic datasets Cora~\cite{yang2016revisiting}, CiteSeer~\cite{yang2016revisiting}, CoraML~\cite{yang2016revisiting}, PubMed~\cite{yang2016revisiting}, Computers~\cite{shchur2018pitfalls}, Photo~\cite{shchur2018pitfalls}, and four heterophily datasets Chameleon~\cite{Pei2020Geom-GCN:}, Squirrel~\cite{Pei2020Geom-GCN:}, Cornell~\cite{Pei2020Geom-GCN:}, and Wisconsin~\cite{Pei2020Geom-GCN:}. 
We consider the scenarios in which model providers will pretrain the GNNs on large-scale graphs, then users utilize GPL on cross-domain graphs. 

\textbf{Defense baseline.} We compare ProGAP with the eight SOTA defense methods.  GNNGuard \cite{gnnguard}, GCN-SVD \cite{gcnsvd}, JaccardGCN~\cite{jaccerd}, HANG-quad \cite{hang}, GADC \cite{gadc}, GPR-GAE \cite{gprgae}, GRAND \cite{grand}, Soft-Median-GDC \cite{zhang2023chasing}. The best performance is highlighted in bold. 

\textbf{Attacks.} For non-adaptive attacks, we select three representative attacks of different types, including meta-learning-based attack (MetaAttack \cite{metatack}), confidence-aware gradient attack (GraD \cite{grad}), and distribution-based attack (Heuristic attack \cite{li2023revisiting}).  For adaptive attacks, we use Greedy-Grad-Descent \cite{adaptive_attack}. 

\subsection{Evaluation of Robustness of ProGAP (RQ1)}
We evaluated our scheme with SOTA defenses under three structural attack scenarios on clean graphs as well as under 25\% and 50\% perturbation rates. \autoref{tab:performance_node_task} presents the results. We draw the following conclusions:

\textbf{The performances of evaluated models remain equally strong when unattacked}, which indicates that all methods possess fundamental graph learning capabilities. While GCN-SVD and GNNGuard exhibit slightly better performance on clean graphs, which suggest limited robustness against adversarial attacks.

\textbf{ProGAP demonstrates substantially stronger robustness and generalization capabilities.} Specifically, our scheme achieves the best performance in 22 out of 28 adversarial settings, outperforming all baseline methods by a clear margin in most cases, particularly at higher perturbation rates. 
Although GADC and GPR-GAE occasionally achieve competitive or even superior results, their performance exhibits instability across different datasets. In contrast, ProGAP maintains stable performance across datasets that vary in structural characteristics.

\textbf{ProGAP demonstrates effective cross-domain generalization capabilities.} Even when pretrained on disparate product sales datasets (Computers and Photo), our scheme continues to achieve SOTA performance, which indicates that our scheme learns general, domain-agnostic representations during the upstream phase rather than domain-specific, overfitted features.

\begin{figure}[t]
    \centering
    \begin{subfigure}{0.23\textwidth}
        \centering
        \includegraphics[width=\linewidth]{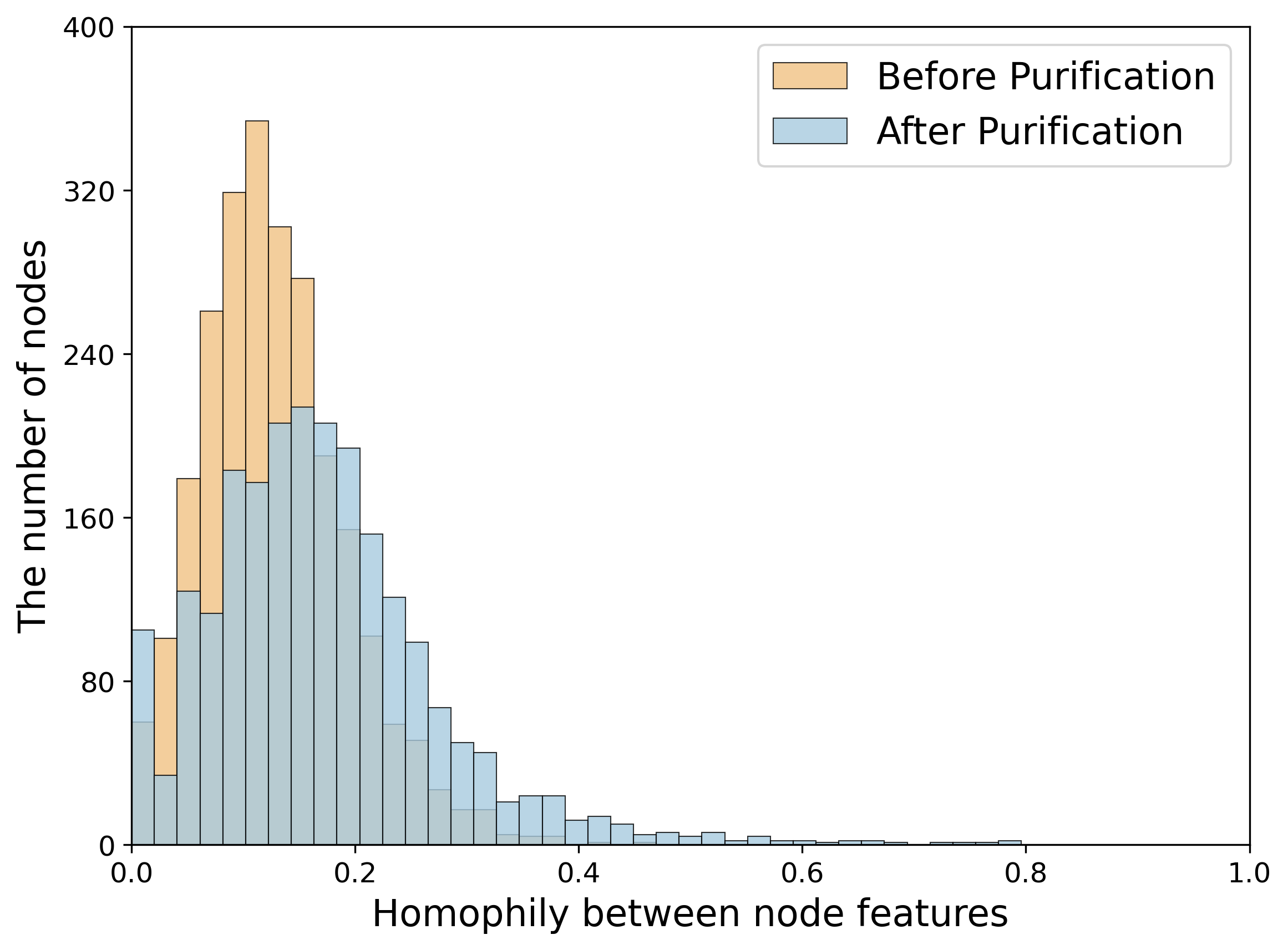}
        \caption{Cora}
        \label{fig:feature-homology-cora}
    \end{subfigure}
    \hfill
    \begin{subfigure}{0.23\textwidth}
        \centering
        \includegraphics[width=\linewidth]{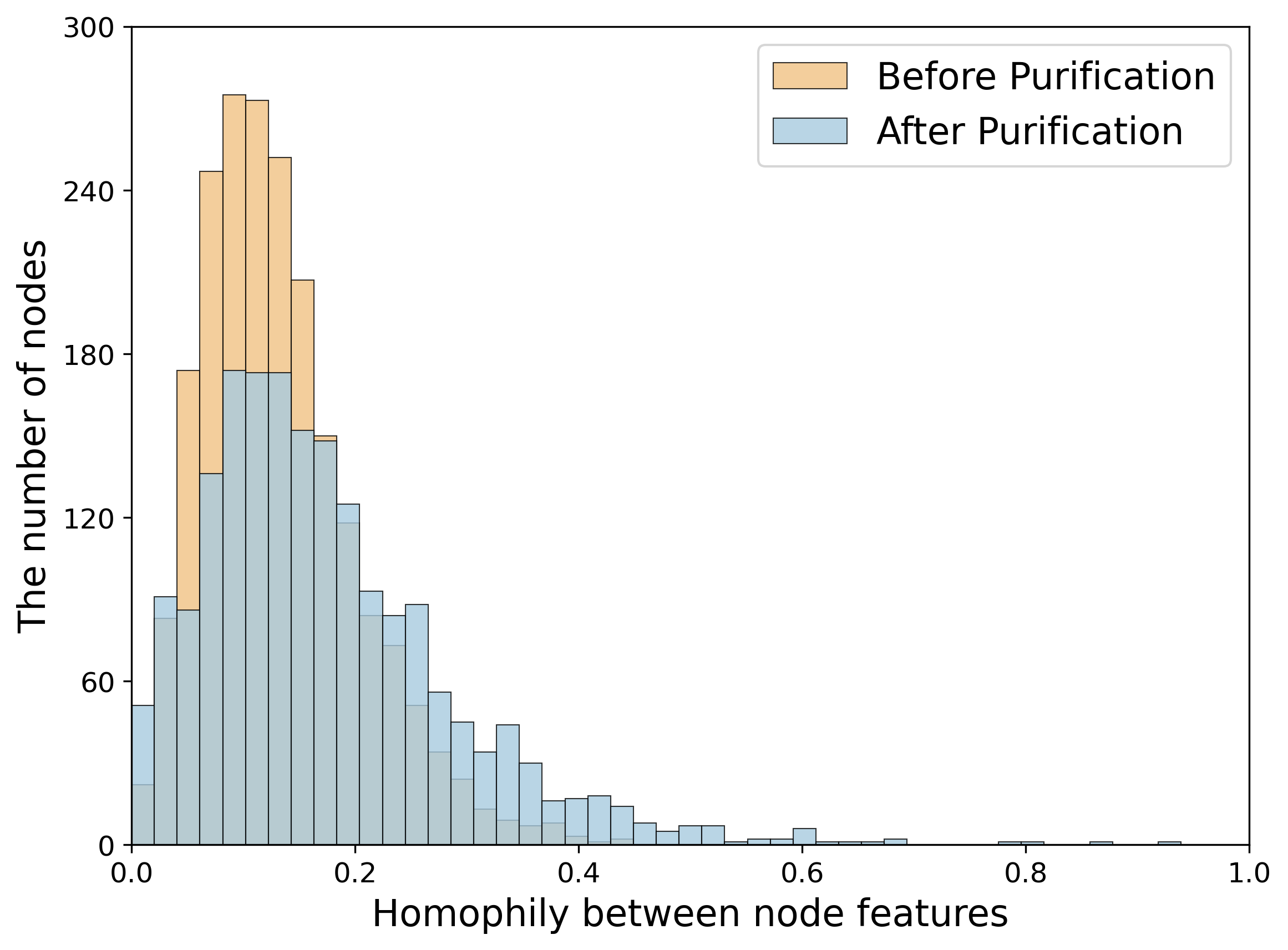}
        \caption{CiteSeer}
        \label{fig:feature-homology-citeseer}
    \end{subfigure}
    \caption{The change of prompted features homology on downstream tasks.}
    \label{fig:similarity1}
    \vskip -3mm
\end{figure}

\begin{figure}[t]
    \centering
    \begin{subfigure}{0.23\textwidth}
        \centering
        \includegraphics[width=\linewidth]{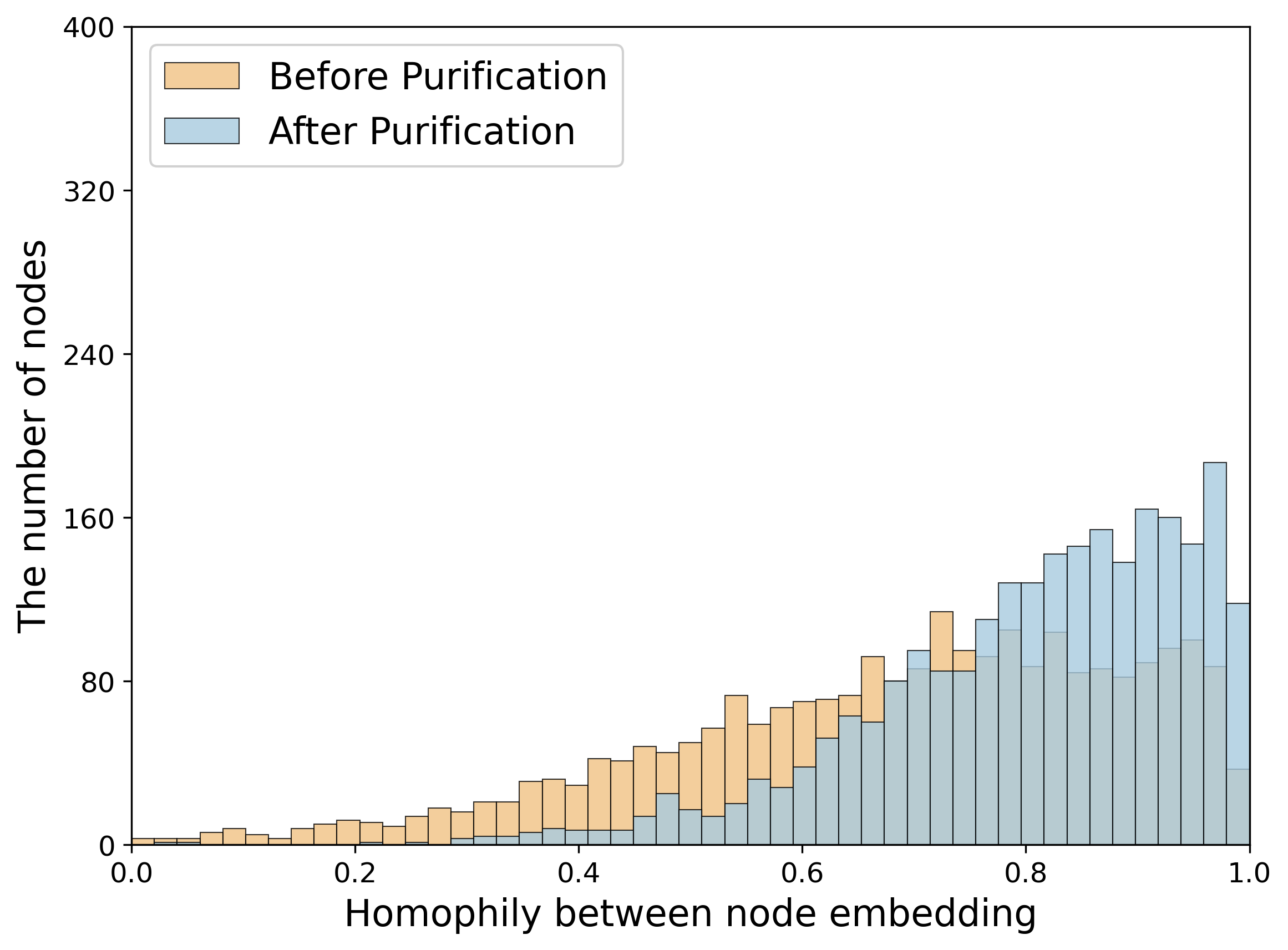}
        \caption{Cora}
        \label{fig:embedding-homology-cora}
    \end{subfigure}
    \hfill
    \begin{subfigure}{0.23\textwidth}
        \centering
        \includegraphics[width=\linewidth]{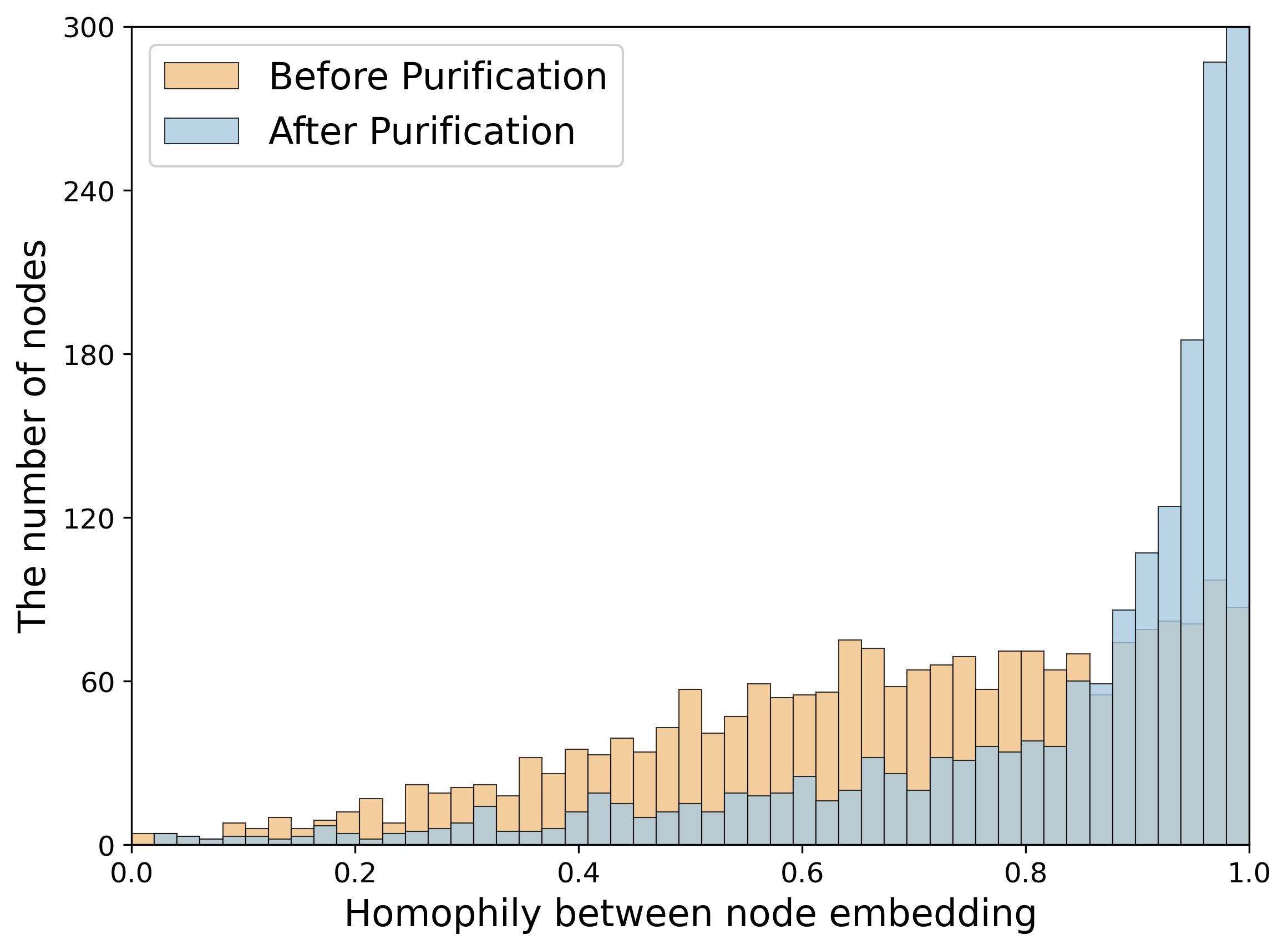}
        \caption{CiteSeer}
        \label{fig:embedding-homology-citeseer}
    \end{subfigure}
    \caption{The change of node embedding homology on downstream tasks.}
    \label{fig:similarity2}
    \vskip -3mm
\end{figure}

\textbf{ProGAP restores structural consistency disrupted by attacks.} The homophily change of node features and node embedding after purification are visualized in \autoref{fig:similarity1} and \autoref{fig:similarity2}. Our scheme substantially improves homophily in both node features and learned embeddings, which demonstrates that our scheme can restore graph properties that are disrupted by adversarial edges.
\begin{figure}[!htbp]
    \centering
    \includegraphics[width=0.45\textwidth]{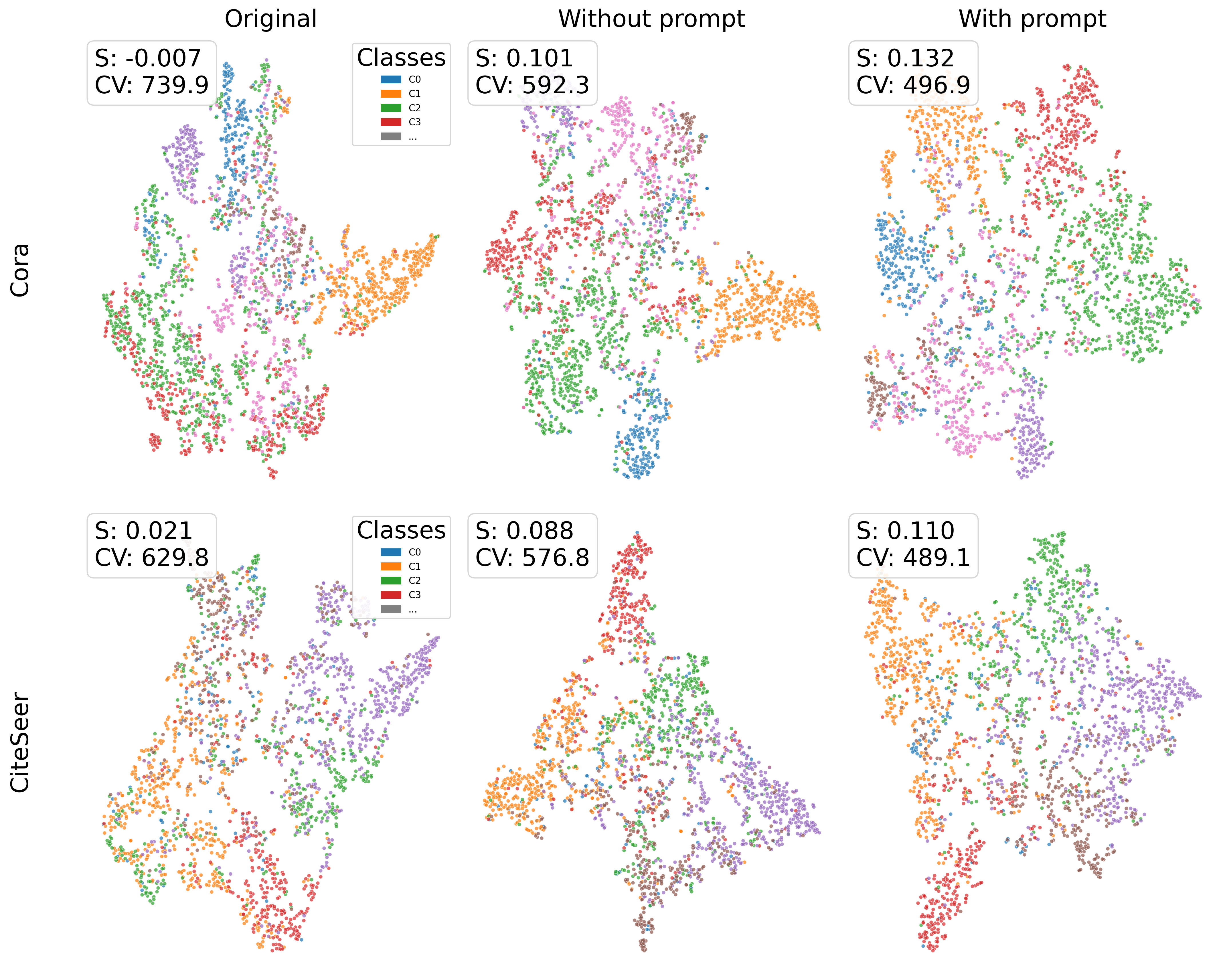}
    \caption{The visualization of node embedding distribution.}
    \label{fig:migration}
    \vskip -3mm
\end{figure}

\subsection{Analysis of the Impact of GPL (RQ2)}
To evaluate whether GPL effectively transfers adversarial knowledge, we compare our scheme against the direct application of the pretrained detector (Without Prompt) under three attack types. Meanwhile, to demonstrate that our vulnerability-aware prompts achieve more robust transfer than general-purpose prompts, we compare our scheme with two representative GPL baselines according to benchmark work~\cite{prog}: GPF and GPF-Plus. As shown in \autoref{fig:prompt}, we draw the following conclusions:

\textbf{GPL improves model robustness across datasets and attacks.} The performance of GPL-migrated pretrained models generally exhibited superior performance compared to directly applied pretrained models. Furthermore, as shown by the visualization of node embedding distributions on CiteSeer and Cora in \autoref{fig:migration}, GPL leads to more stable decision boundaries with improved inter-class separation~(S) and reduced intra-class variance~(CV).

\textbf{Compared to generic prompts, our vulnerability-aware prompts demonstrate superior robust cross-domain transfer.} This improvement stems from our explicit modeling of structural vulnerabilities: while GPF~/~GPF-Plus offer generic feature space adaptation, our prompts specifically target protecting biased nodes, which are the primary targets of adversarial attacks. 

\begin{table}[h]
\centering
\caption{Total time (in seconds) consumed by different defense methods.}
\label{tab:effciency}
\setlength{\tabcolsep}{5pt}
\begin{tabular}{cccccc}
\toprule
Defense & Cora & CiteSeer & CoraML & PubMed \\
\midrule
GCN-SVD     & 4.20  & 4.96 & 6.05 & 169.53 \\
HANG-quad     & 10.33 & 11.58 & 9.37 & 22.85 \\
GADC & 6.93 & 7.13 & 7.02 & 13.10 \\
GPR-GAE & 29.47 & 25.84 & 35.85 & 53.46 \\
ProGAP & 4.09  & 5.15 & 5.93 & 10.11 \\
\bottomrule
\end{tabular}
\end{table}
\begin{table*}[h]
\caption{The accuracy ($\%\ \pm\ \sigma$) under adaptive attacks. OOM indicates out-of-memory.}
\label{tab:adaptive}
\centering
\setlength{\tabcolsep}{3.4pt} 
\begin{tabular}{lcccccccc}
\toprule
Model 
& \multicolumn{4}{c}{Evasion Attack} 
& \multicolumn{4}{c}{Poisoned Attack} \\
\cmidrule(r){2-5} \cmidrule(l){6-9}
& Cora & CiteSeer & PubMed & CoraML 
& Cora & CiteSeer & PubMed & CoraML \\
\midrule
GCN       & 59.03$\pm$0.64 & 45.35$\pm$0.86 & 82.44$\pm$0.04 & 64.76$\pm$0.20 & 59.67$\pm$0.62 & 45.64$\pm$0.73 & 82.54$\pm$0.09 & 64.77$\pm$0.19 \\
GCN-SVD   & 56.81$\pm$3.41 & 45.71$\pm$1.81 & 79.01$\pm$0.09 &65.84$\pm$0.27 & 56.73$\pm$0.73 & 45.88$\pm$1.67 & 79.47$\pm$0.08 & 67.41$\pm$0.31\\
GNNGuard  & 61.47$\pm$0.56 & 48.03$\pm$0.60 & OOM & 66.50$\pm$0.44& 62.06$\pm$0.34 & 48.84$\pm$1.02 & OOM & 66.71$\pm$0.33 \\
GRAND     & 51.74$\pm$1.27 & 47.13$\pm$1.96 & 81.05$\pm$0.19 &27.80$\pm$0.02& 45.69$\pm$3.01 & 49.45$\pm$3.73 & 80.74$\pm$1.03 &27.80$\pm$0.02 \\
Soft-Median-GDC   & 63.15$\pm$6.08 & 52.93$\pm$8.31 & OOM &61.16$\pm$4.73& 63.74$\pm$3.03 & 59.82$\pm$2.52 & OOM & 70.43$\pm$1.85\\
\midrule
ProGAP    & \textbf{71.49}$\pm$\textbf{1.36} & \textbf{65.97}$\pm$\textbf{1.56} & \textbf{84.00}$\pm$\textbf{0.24} & \textbf{73.85}$\pm$\textbf{1.25}
          & \textbf{72.14}$\pm$\textbf{0.89} & \textbf{67.69}$\pm$\textbf{0.54} & \textbf{84.64}$\pm$\textbf{0.24} & \textbf{75.60}$\pm$\textbf{2.39} \\
\bottomrule
\end{tabular}
\vskip -3mm
\end{table*}
\begin{table*}[h]
\centering
\caption{The accuracy ($\%\ \pm\ \sigma$) of ablation study on $D_{t}$, $D_{s}$, attention mechanism, pseudo labels and prompts ($p_d$, $p_s$).}
\label{tab:ablation}
\setlength{\tabcolsep}{5.5pt} % 减小列间距，默认是6pt
\begin{tabular}{lcccccccc}
\hline
Variants & \multicolumn{2}{c}{Cora} & \multicolumn{2}{c}{CiteSeer} & \multicolumn{2}{c}{CoraML} & \multicolumn{2}{c}{PubMed} \\
\cmidrule(lr){2-3} \cmidrule(lr){4-5} \cmidrule(lr){6-7} \cmidrule(lr){8-9}
 & G-25 & M-25 & G-25 & M-25 & G-25 & M-25 & G-25 & M-25 \\
\hline
ProGAP & \textbf{76.94}$\pm$\textbf{1.39} & \textbf{77.41}$\pm$\textbf{0.89} & 70.05$\pm$1.49 & \textbf{70.92}$\pm$\textbf{1.22} & \textbf{79.12}$\pm$\textbf{0.51} & \textbf{78.62}$\pm$\textbf{1.21} & 82.22$\pm$0.70 & 83.64$\pm$0.51 \\
w/o $D_{t}$ & 60.50$\pm$3.57 & 62.83$\pm$3.30 & 62.86$\pm$0.95 & 62.91$\pm$0.44 & 59.80$\pm$3.44  & 67.74$\pm$3.48& 78.50$\pm$0.77 & 78.45$\pm$0.73 \\
w/o $D_{s}$ & 65.66$\pm$0.50 & 65.71$\pm$1.08 & 64.09$\pm$0.73& 64.72$\pm$2.14 & 57.07$\pm$2.56 & 65.92$\pm$3.75 & 80.70$\pm$0.58 & 79.39$\pm$0.47 \\
w/o attn. & 75.40$\pm$1.27 & 75.57$\pm$1.11 & 69.90$\pm$0.86 & 70.50$\pm$1.30 & 78.59$\pm$0.45 & 77.40$\pm$1.18 & \textbf{82.72}$\pm$\textbf{0.75} & \textbf{83.77}$\pm$\textbf{0.38} \\
w/o pl  & 76.13$\pm$1.74 & 77.08$\pm$1.23 & \textbf{70.09}$\pm$\textbf{1.09} & 70.23$\pm$1.03 & 78.90$\pm$0.41 & 78.08$\pm$1.35 & 81.95$\pm$0.77  & 83.26$\pm$0.33 \\
w/o $p_{s}$ & 69.86$\pm$4.25 & 75.07$\pm$0.94 & 67.44$\pm$0.96 & 68.52$\pm$1.23 & 73.81$\pm$1.04 & 76.06$\pm$0.54 & 81.46$\pm$0.52  & 82.40$\pm$0.66 \\
w/o $p_{d}$ & 75.14$\pm$1.21 & 74.55$\pm$3.05 & 69.59$\pm$0.64 & 70.86$\pm$0.59 & 76.29$\pm$1.45 & 73.71$\pm$2.55 & 82.01$\pm$0.92  & 83.26$\pm$0.33 \\
\hline
\end{tabular}
\vskip -2mm
\end{table*}
\subsection{Study for Time Consumption (RQ3)}\label{sec:efficiency}
To quantify defense efficiency, we compared the overall runtime of ProGAP with several competitive baselines, with results shown in \autoref{tab:effciency}. Experimental results demonstrate that \textbf{our scheme achieves significantly higher defense effectiveness} while maintaining robustness and generalization capabilities. This efficiency gain primarily stems from two carefully designed features in our scheme: First, the prompt learning mechanism reduces the number of parameters requiring model updates to a single token, substantially lowering computational overhead. Second, the transferable edge detector eliminates the need to perform both perturbation generation and classification stages separately on each graph, further optimizing the overall workflow. 
\subsection{Evaluate Robustness under Adaptive Attacks (RQ4)}
We conducted experiments under adaptive poisoned attacks and evasion attacks targeting classical defenses. As shown in \autoref{tab:adaptive}, ProGAP significantly outperforms baseline models in all scenarios. This demonstrates that even when attackers obtain complete information about our defense mechanism, which includes model architecture and training procedures, our scheme can still effectively reconstruct the adjacency matrix and provide robust defense.

\subsection{Study on the Impact of Components (RQ5)}
To disentangle the contribution of each component in our scheme, we perform six ablation studies: w/o $D_{t}$ is a variant that does not use $D_{t}$, w/o $D_{s}$ is a variant that does not use $D_{s}$, w/o attn. is a variant that does not use Attention. w/o pl is a variant that does not use pseudo labels. w/o $p_{s}$ is a variant that does not use the similarity vulnerability prompt. w/o $p_{d}$ is a variant that does not use the degree vulnerability prompt. The result is shown in \autoref{tab:ablation}, we observe following key points:

\textbf{Semantic and structural modeling jointly determine the performance.} Removing either the semantic $D_{s}$ or the topological $D_{t}$ causes substantial degradation, which indicates that neither type of information is sufficient on its own. The removal of prompts ($p_d, p_s$) also results in an accuracy drop, which is designed for these considerations. 

\textbf{Attention mechanism contributes to the effective fusion.} Removing attention results in moderate performance drops. While the drops are smaller than removing individual modules, the attention mechanism balances structural and semantic signals for optimal edge scoring.

\textbf{The contribution of pseudo labels is relatively limited.} Excluding pseudo labels only causes a slight drop, which shows that the core detection capability relies primarily on the perturbation-capture edge detector and vulnerability-aware prompts rather than on extra supervision.
\subsection{Evaluation of Cross-homophily Transferability~(RQ6)}
\begin{figure}[t]
    \centering
    \begin{subfigure}{0.235\textwidth}
        \centering
        \includegraphics[width=\linewidth]{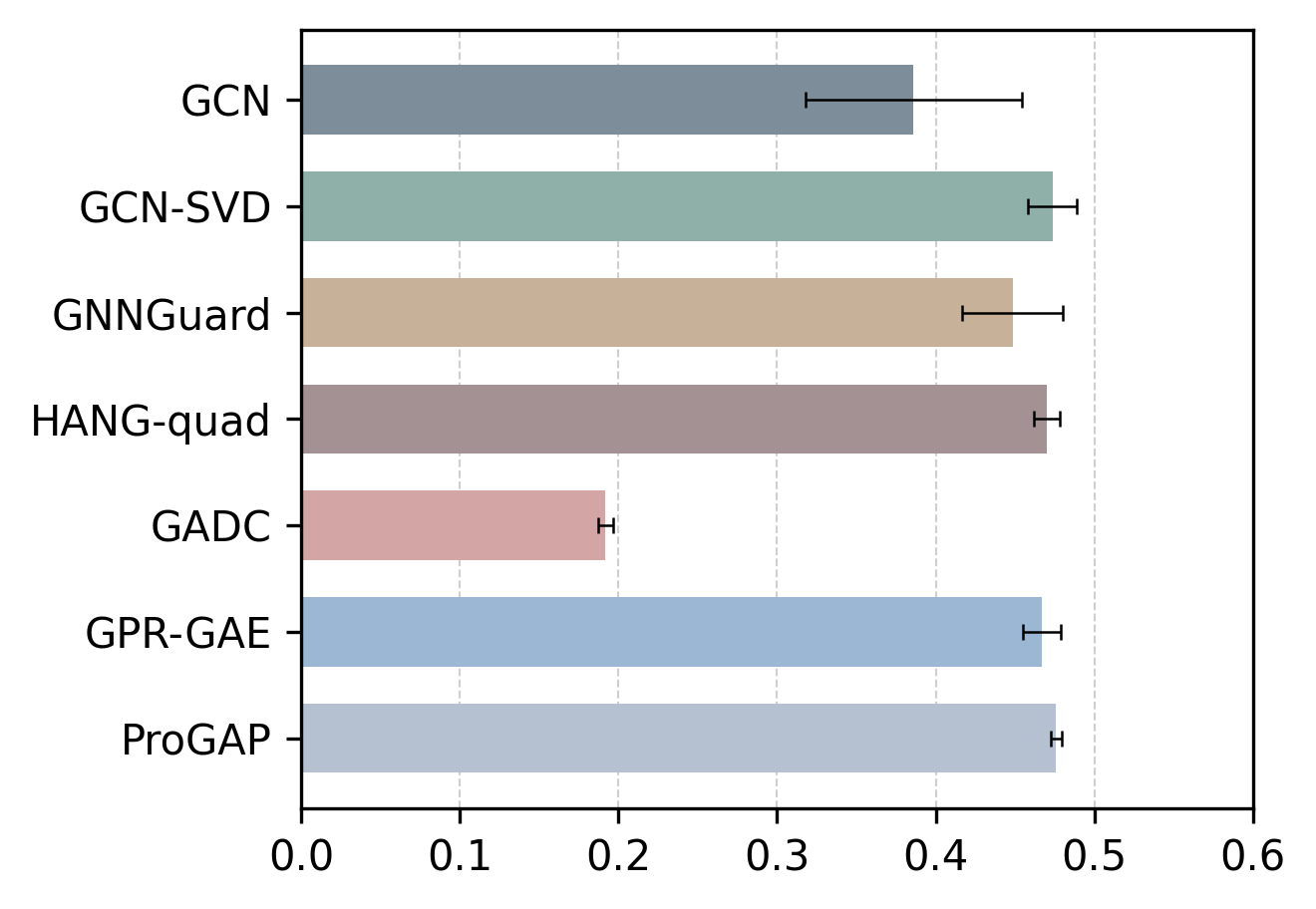}
        \caption{Chameleon~(M-10)}
        \label{fig:homo-chameleon-10}
    \end{subfigure}
    \hfill
    \begin{subfigure}{0.235\textwidth}
        \centering
        \includegraphics[width=\linewidth]{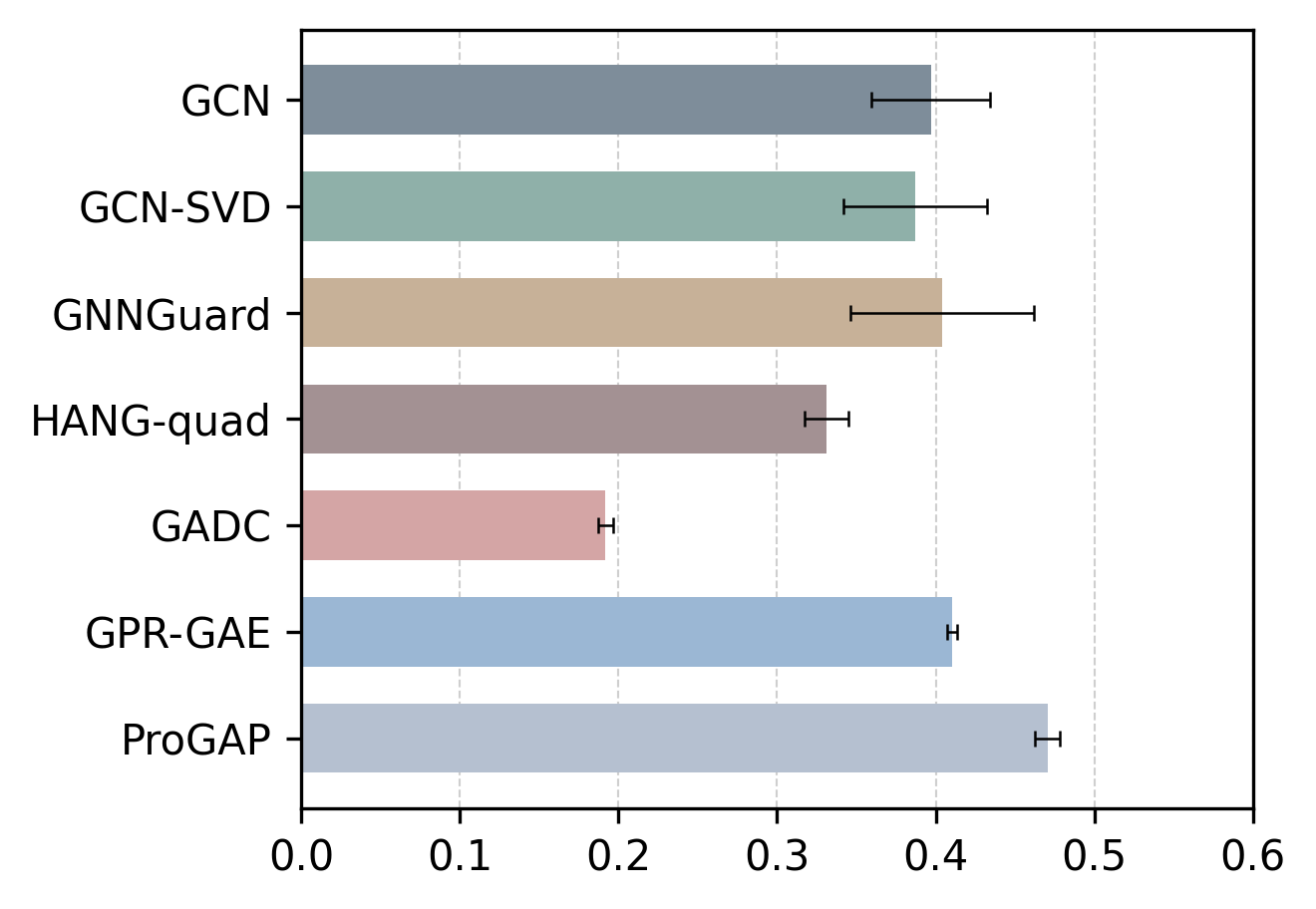}
        \caption{Chameleon~(M-25)}
        \label{fig:homo-chameleon-25}
    \end{subfigure}

    \begin{subfigure}{0.235\textwidth}
        \centering
        \includegraphics[width=\linewidth]{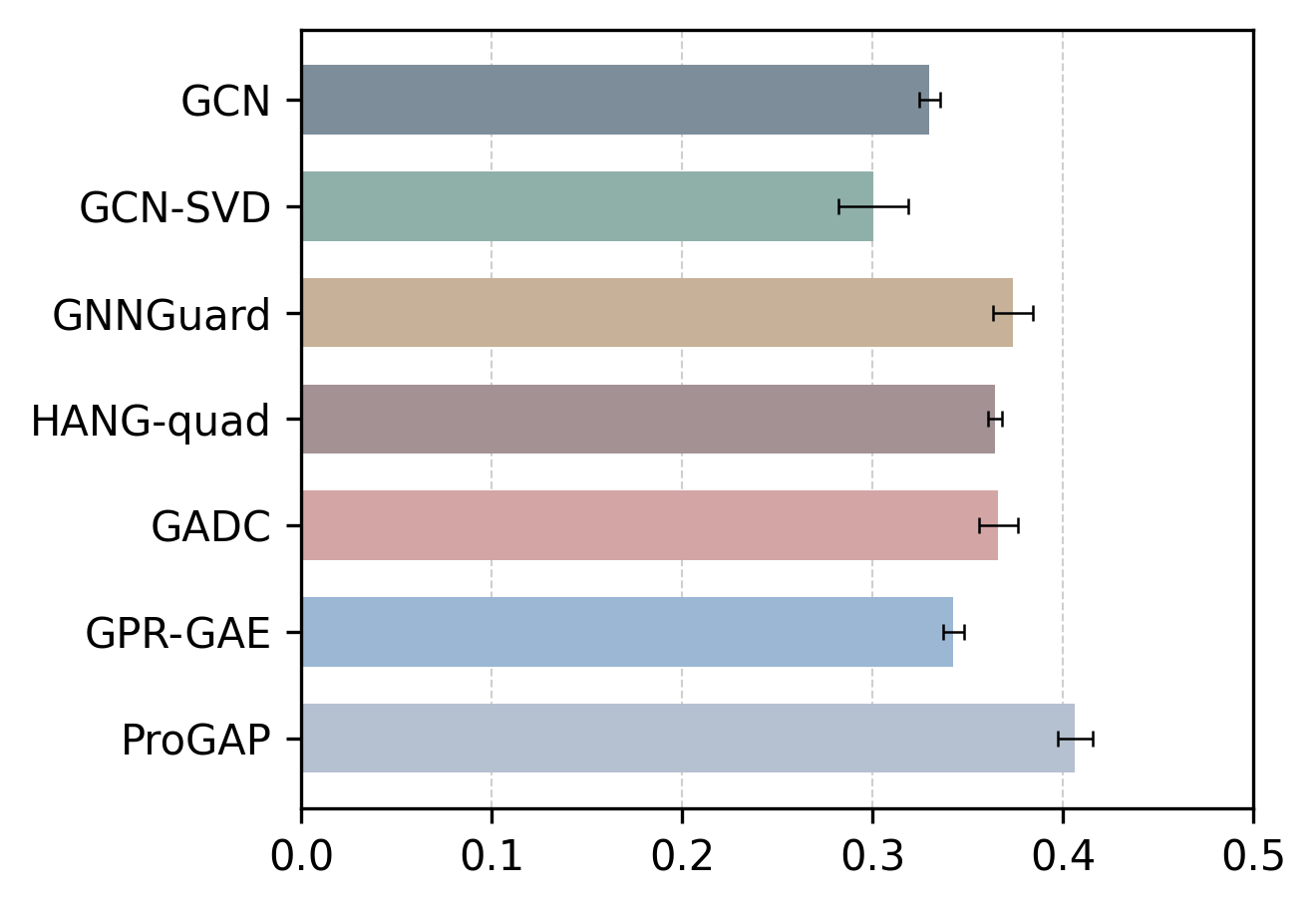}
        \caption{Squirrel~(M-10)}
        \label{fig:homo-squirrel-10}
    \end{subfigure}
    \hfill
    \begin{subfigure}{0.235\textwidth}
        \centering
        \includegraphics[width=\linewidth]{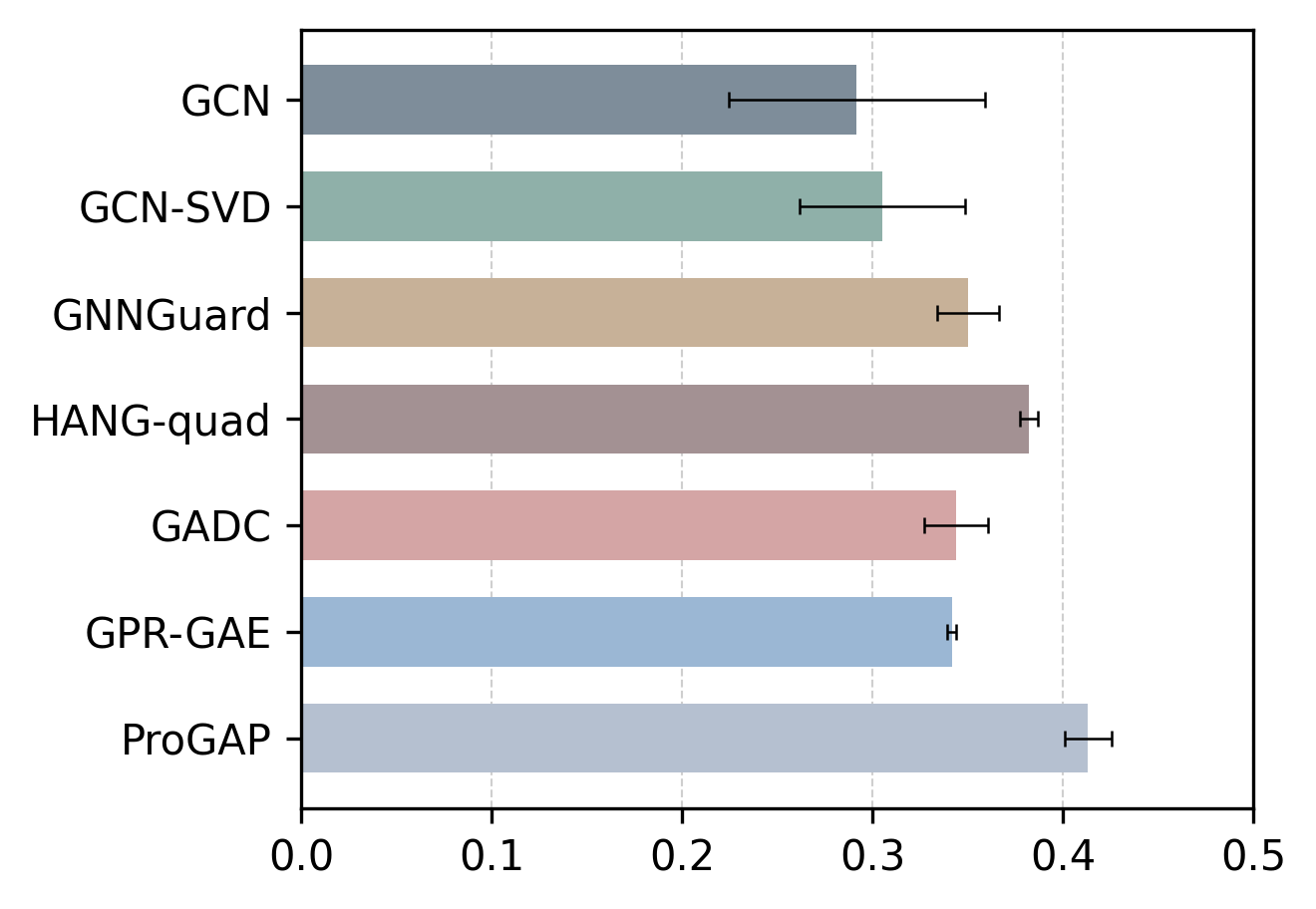}
        \caption{Squirrel~(M-25)}
        \label{fig:homo-squirrel-25}
    \end{subfigure}

    \caption{The accuracy comparison between our scheme and baselines on heterophily graphs under clean and Mettack settings.}
    \vskip -6.5mm
    \label{fig:homo}
\end{figure}
To evaluate ProGAP’s generalization ability in scenarios with significant structural differences, we transferred a model pretrained on a homophily reference network to two heterophily datasets: Chameleon and Squirrel. These target domains exhibit heterogeneity, which represents a fundamental shift from the structural assumptions of the source domain and reflects differences in graph structure distributions. The evaluation is conducted under the Mettack adversarial attack with perturbation rates of 10\% and 25\%. As shown in \autoref{fig:homo}, the results indicate that across various perturbation strengths, ProGAP achieves SOTA performance compared to baseline defense methods. This superior performance is primarily attributed to the synergy between the perturbation-capture edge detector, which decouples and captures universal adversarial patterns, and the vulnerability-aware prompts that provide adaptive defensive guidance for biased nodes to mitigate distribution shifts.

\section{Conclusion}
In this study, we propose a transferable graph purification scheme, named ProGAP, which addresses the limitation of existing adversarial purification defense methods being confined to a single domain. Our comprehensive research leads to the following conclusions: (1) lightweight prompts can be served as transferable carriers of adversarial topological knowledge across multimedia domains; (2) extensive experiments show our scheme outperforms SOTA techniques across diverse attack scenarios while resisting adaptive attacks; (3) vulnerability-aware prompts achieve superior robust transfer compared to generic prompt by explicitly targeting structurally and semantically vulnerable nodes, given its resemblance to multimedia learning scenarios; and (4) by freezing the pretrained edge detector and optimizing only lightweight prompts, our scheme improves defense efficiency. In the future, we will explore the application of our scheme in more complex and dynamic multimedia scenarios.

\section{Acknowledgements}
This work was supported in part by the National Natural Science Foundation of China under Grant 62402106, in part by the Natural Science Foundation of Jiangsu Province of China under Grant BK20241272, in part by the “Zhishan” Scholars Programs of Southeast University, Grant 2242026RCB0031, and in part by the Start-Up Research Fund of Southeast University under Grant RF1028623129.
\bibliographystyle{ACM-Reference-Format}
\bibliography{sample-base}

\end{document}